# Analytical Techniques to Support Hospital Case Mix Planning

Robert L Burdett[1,3], r.burdett@qut.edu.au (Corresponding Author)
Paul Corry[1], p.corry@qut.edu.au
David Cook[2], d.cook@uq.edu.au
Prasad Yarlagadda[3], y.prasad@qut.edu.au

[1] School of Mathematical Sciences, Queensland University of Technology, GPO Box 2434, 2 George Street Brisbane Qld 4000, Australia
[2] Princess Alexandra Hospital, 2 Ipswich Rd, Woolloongabba, Brisbane, Qld 4102, Australia
[3] School of Mechanical, Medical & Process Engineering, Queensland University of Technology, GPO Box 2434, 2 George Street Brisbane Qld 4000, Australia

**Abstract –** This article introduces analytical techniques and a decision support tool to support capacity assessment and case mix planning (CMP) approaches previously created for hospitals. First, an optimization model is proposed to analyse the impact of making a change to an existing case mix. This model identifies how other patient types should be altered proportionately to the changing levels of hospital resource availability. Then we propose multi-objective decision-making techniques to compare and critique competing case mix solutions obtained. The proposed techniques are embedded seamlessly within an Excel Visual Basic for Applications (VBA) personal decision support tool (PDST), for performing informative quantitative assessments of hospital capacity. The PDST reports informative metrics of difference and reports the impact of case mix modifications on the other types of patient present. The techniques developed in this article provide a bridge between theory and practice that is currently missing and provides further situational awareness around hospital capacity.

## 1. Introduction

This article considers case mix planning (CMP) in hospitals. This is the problem of identifying a patient cohort (a.k.a., case mix) with a specific set of features deemed desirable or ideal. Identifying the ideal composition and number of patients to be treated, however, is not straight-forward and is quite nuanced. A variety of challenges make this task challenging (Hof et. al. (2017)). First, there are many different alternative case mixes that can be selected. Some case mixes are favourable for some patient type and unfavourable for others. Second, the term "ideal" is subjective and can mean different things in a practical setting. A case mix may be sought that is most equitable, for instance in the allocation and usage of hospital resources. A case mix may also be sought that is most economical or financially viable to treat. From a utilization and output-oriented perspective, a maximal cohort may also be sought. That cohort results in the greatest number of patients treated over time. A maximal cohort saturates the resources of the hospital and is a measure of the hospitals' capacity. Identifying a case mix that meets or exceeds specified demands or targets is also of significant interest. Last, quality of care is especially important in all the variants mentioned.

     CMP is often made more complex by either a lack of precise information, a high volume of unrefined empirical data, and stochastic parameters (Burdett et. al. 2022a). The presence of stochastic treatment durations and length of stay make it difficult to exactly ascertain how long each hospital resource would be utilized over a specified time. The categorisation of patient type is also a vital ingredient in CMP, but the approach taken is rarely straightforward, often ad-hoc and politically sensitive. There are many different types of illness, and many medical and surgical treatments. Categorising patients into finite groups with common features is subjective, and any choice that is made can easily skew the results. Patients may be grouped by specialty, diagnostic resource group (DRG), international classification of disease (ICD), similar resource consumption and treatment duration, or other strategies (Burdett et. al. 2017). In CMP there is a choice to include, surgical, medical, or acute patients, and to consider a hospital holistically or piecemeal. It is worth mentioning

that CMP may also refer to the allocation of available operating theatre time amongst different surgical specialties and surgical patient types.

CMP is an active research area and has been considered on numerous occasions. A variety of quantitative methods have been applied for CMP in hospitals. In most cases, the activity is performed parametrically without observation of actual hospital operations. For instance, an analysis is performed relative to key hospital parameters and other geographic details, like hospital layout and configuration. The articles by Burdett and Kozan (2016), Burdett et. al. (2017), Fulshof et. al. (2017), McRae et. al. (2018), McRae and Brunner (2019), Mahmoudzadeh et.al. (2020) demonstrate the current state of the art. Their focus is predominantly the identification of a favourable case mix, subject to resource availability constraints. Our review of the literature suggests that CMP across multiple hospitals has yet to be comprehensively addressed and stochastic assessments are also quite new (Burdett et. al. (2022b), Mahmoudzadeh et.al. (2020)). The development of appropriate decision support tools is severely lacking, and only the recent work by Burdett et. al. (2022b) considers how hospital planners, executives, and layperson, could perform CMP. Overall, there is a lack of methodological support for health care professionals in their decision making, encompassing the provision and allocation of finite hospital resources (Bruggemann et.al. 2021).

From a general viewpoint it's evident that strategic, tactical, and operational level decision problems are abundant in health care, and much research is being performed in this field. Operational problems are popular, and some examples include Chen et. al. (2015), Spratt and Kozan (2016), Liu et. al. (2019), Hawkinson et.al. (2018). Providing advanced predictive analytics is an important goal in the healthcare industry (Kreuger (2018)) and the development of planning software, that is user-friendly, intuitive, and extensible is challenging, but vital. Krueger (2018) reports the emerging use of operational tools and science in health care. However, we should also acknowledge that there has been restricted uptake and development of advanced hospital planning tools and software to date (Burdett et. al. 2022b). Trust in the Operations Research (OR) team building the tool is reported as most vital to success (Pagel et.al. 2017). Predicting patient flows is a vital task and provides information that can be used in hospital case mix planning and capacity assessment. Resta et.al. (2018) developed artificial neural networks to classify, cluster and predict patient flows. They used their self-organizing map to gain a deeper understanding of patient flows in an emergency department in Italy. The potential for capacity planning research in nursing and other community care facilities is also emerging. Frail and elderly patients are highly vulnerable and are often a cause of bed blocking within hospitals due to the lack of or unsuitable community care resources available (Williams et.al. 2021). It is therefore important for research to focus on hospitals and how they feed into community care services. Multi-criteria decision making (MCDM) has increased greatly in healthcare in recent years (Chalgham et.al., 2019). Chalgham et.al. (2019) proposed MCDM methods to improve in-patient flows from an emergency department. Malik et.al. (2015) formulated and solved a bi-objective aggregate capacity planning problem for operating theatres. The considered the minimization of elective patients waiting for treatment and minimization of healthcare expenditure. Both of the aforesaid articles, however, did not consider an entire hospital and the effect on downstream resources, likes beds. Emergency management of health care facilities in the event of various disasters and crises is also an active area for which this articles' approach is applicable. High level emergencies have serious consequences on hospital activities (Chen, Guinet and Ruiz, 2015). Chen, Guinet and Ruiz (2015) considered hospital evacuation planning and developed a simulation model to analyse the process.

**Research Aims.** In this article some supporting techniques are proposed to assist with hospital CMP. The supporting techniques we have developed are deemed necessary for a variety of reasons and do not currently exist in the literature. These will be discussed in due course. First, we consider how a specific change to an existing patient case mix impact resource availability for other patient type. Specifically, we would like to know which patient types are affected, and by how much are they affected. For instance, can more, or less patients be treated? We think this functionality is necessary because a patient case mix that meets all goals and expectations is unlikely to be found immediately.

There are many qualitative factors and stakeholders, and some slight adjustment of the mix would typically be necessary. To identify the effect, we propose a mathematical programming model. The capability to analyse changes of this nature and to immediately see the effect will improve situational awareness for hospital executives, managers, and planners who are weighing up the pros and cons of treating more, or less patients of a given type. This type of enquiry in theory facilitates a broader sensitivity analysis that can provide significant insights into how a hospital can be operated, and who they can treat. Previously, the goal of CMP was just to identify a single favourable case mix solution, or to provide a set of non-dominated case mix solutions, with no understanding of the difference between those.

Second, we consider how to compare different patient case mix to better facilitate CMP in a multi-criterion setting (Burdett and Kozan (2016)). In a multi-criterion setting, it is necessary to generate a Pareto frontier, which is set of alternatively optimal solutions. How to choose one solution over another is very unclear. We would like to decide rationally, and to have a consistent framework to substantiate any choice. To this end, analytical aids are needed.

The techniques developed in this article motivate the development of a decision support tool for hospital managers and planners to use. Without such a tool, it is hard to imagine, that the techniques advocated in this article could be applied in a practical setting. It seems unlikely that any hospital or health care provider would have the time and resources to create a tool that does what this paper suggests, so it falls to us to prove the concept. We test how the new techniques can be embedded within a personal decision support tool to perform the described CMP analyses.

The format of this article is as follows. In Section 2 the details of the analytical methods and framework is provided, commencing with an outline of key technical details. The specification, capabilities and graphical user interfaces are then presented in Section 3. Design strategies employed during development are also provided and examples of how the GUI are used to perform various assessments is shown. Conclusions and final remarks are given in Section 4. Broader issues including potential further extensions to the software are also discussed.

## 2. Methodology

### 2.1. Background Details.

The formal details and notation relevant to later modelling are now described. To avoid confusion, it is important to note that in *some limited circumstances we reuse some symbols and differentiate parameters by their index*. This abuse of notation is beneficial as fewer symbols need to be introduced and understood.

For CMP we first recognise that are limited operating theatres, wards, and beds in a hospital. Operating theatres and wards consist of beds where treatments and care are provided. They are viewed as distinct precincts (a.k.a., zones) for health care. The set of hospital precincts are denoted by $W$, and for each zone $w \in W$ the number of treatment spaces contained within is denoted $b_w$. Therefore, there are $\sum_{w \in W} b_w$ treatment spaces in total.

It is assumed that there is a clearly defined set of patient types (a.k.a., groups) that are treatable. For each patient type $g \in G$, there is a set of patient sub-types denoted $P_g$. The full set of patient sub types is hence denoted $GP = \{(g,p) | g \in G, p \in P_g\}$. For each element of $GP$ there is a specific clinical pathway, denoted $A_{(g,p)}$. The pathway is essentially a set of medical and surgical activities, each with a planned treatment time and location. As such, we can write $A_{(g,p)} = \{(g,p,k) | (g,p) \in GP\}$. The activity time measured in either hours or minutes is $t_{(g,p,k)}$ and the set of candidate locations for the activity is denoted $l_{(g,p,k)}$. The set of all activities is $A = \bigcup_{(g,p) \in GP} A_{(g,p)}$ and those that can be performed in zone $w$ is $A_w = \{a \in A | w \in l_a\}$.

The number of patients of type $g$ treated in hospital $h$ and the number of patients of sub-type $p$ is to be determined. These constitute the case mix and sub mix of the hospital. They are respectively denoted $n_g^1$ and $n_{g,p}^2$. The total number of patients treated is $\mathbb{N}$. It is worth noting that, $\mathbb{N} = \sum_g n_{h,g}^1 \ \forall g \in G, n_{h,g}^1 = \sum_{p \in P_g}(n_{h,g,p}^2) \ \forall g \in G, \forall p \in P_g$.

Within each precinct, the treatment spaces are available for a specified number of hours per period. The time availability of $w \in W$ is $T_w$. The number of activities of type $a = (g, p, k)$ performed in precinct $w$ is denoted $\alpha_{a,w}$. It is worth noting that $\alpha_{a,w} = 0 \ \forall w \in W \setminus l_a$, i.e., activities cannot be performed outside of their candidate locations. The decision captured by $\alpha_{a,w}$ for all $a$ and $w$ is referred to as the allocation. It is inherently linked to the decisions $n_g^1$ and $n_{g,p}^2$.

A CMP model has been proposed previously and will be used as the basis upon which this articles' quantitative techniques are based. The CMP model is listed in Appendix A for interested readers. The CMP model determines maximum number of patients that treatable over a specified period, subject to limited time availabilities of operating theatres, wards and beds, and a required case mix.

## 2.2. Case Mix Alterations (CMA)

For a given case mix $n^1$, the effect of a change in the number of patients of a single type $g^*$, is worth understanding. If patient type $g^*$ is increased and anchored, then it may be necessary to decrease several other patient types, as less capacity is available for them. If patient type $g^*$ is decreased and anchored, then the opposite occurs, i.e., more capacity is available for the others. The following steps are required to facilitate a query such as described:

Step 1. Select a current case mix $n^1$, and designate targets $\hat{n}_g^1 = n_g^1$.
Step 2. Select a change of $\delta$ units in one patient type $g^*$, i.e., $(\delta, g^*) | 0 \leq \hat{n}_{g^*}^1 + \delta \leq \overline{n}_{g^*}^1 0$ where $\delta \neq 0$ and $\overline{n}_g^1$ is the upper bound on the number of patients of type $g$ that are treatable.
Step 3. If $\delta > 0$ then the difference $\|n^1 - \hat{n}^1\|$ should be minimized, else if $\delta < 0$ then the difference should maximized, where $\|n^1 - \hat{n}^1\| = \left(\sum_{g \in G}(\gamma_g)^2\right)^{0.5}$ and $\gamma_g = \hat{n}_g^1 - n_g^1$ or $\gamma_g = \frac{\hat{n}_g^1 - n_g^1}{\overline{n}_g^1}$.
Step 4. Solve the optimization model below to obtain the case mix $n^1$. Report the differences $\gamma_g$.

$$\text{Minimize } Z = sgn(\delta) \times \sum_{g|g \neq g^*}(\gamma_g)^2 \text{ where } sgn(\delta) = \delta/|\delta| \tag{1}$$

Subject to:
$$n_{g^*}^1 = \hat{n}_{g^*}^1 + \delta \quad \text{[Forced number of type } g^*\text{]} \tag{2}$$
$$\gamma_{g^*} = 0 \text{ and } \gamma_g = sgn(\delta) \times (\hat{n}_g^1 - n_g^1)/\overline{n}_g^1 \ \forall g \in G | g \neq g^* \quad \text{[Scaled difference]} \tag{3}$$
$$sgn(\delta) \times n_g^1 \leq sgn(\delta) \times \hat{n}_g^1 \ \forall g \in G | g \neq g^* \text{ [Forced increase or decrease]} \tag{4}$$
$$\gamma_g \geq 0 \ \forall g \in G \quad \text{[Positive scaled difference]} \tag{5}$$
+ Constraint (A2) - (A10) excluding (A9) from Appendix A

This model is hereby denoted CMA-SSQ. Scaling is introduced in constraint (3) to manage the different orders of magnitude that may occur. As a result, differences can be compared more objectively. Without scaling, some types will have more, or less increase (decrease). Equation (3) also ensures that $\gamma_g$ takes positive values in both situations. For instance, $\forall g \in G | g \neq g^*$:

$$\gamma_g = \frac{\hat{n}_g^1 - n_g^1}{\overline{n}_g^1} \text{ if } \delta > 0 \quad \text{and } \gamma_g = \frac{n_g^1 - \hat{n}_g^1}{\overline{n}_g^1} \text{ if } \delta < 0$$

With regards to constraint (4), if $\delta < 0$, then there must be an increase, i.e., $n_g^1 \geq \hat{n}_g^1$. If $\delta > 0$, then there must be a decrease, i.e., $n_g^1 \leq \hat{n}_g^1$. Constraint (5) ensures that a patient type is not increased when it should be decreased or vice versa.

The CMA-SSQ model has a sum of squares objective, and this is a quadratic function. As such, the model can be solved using Quadratic Programming (QP) techniques. However, when $\delta < 0$, we find the model to be non-convex (i.e., the hessian is not positive semi-definite), and QP is not suitable. Separable Programming (SP) techniques, however, may be applied. This requires the objective to be expanded as in (6), and the term $\left(n_g^1/\overline{n}_g^1\right)^2$ to be approximated by a piecewise linear function, i.e., $\left(\frac{n_g^1}{\overline{n}_g^1}\right)^2 \approx y_g$ where $y_g = \text{PWL}(n_g^1/\overline{n}_g^1, f, b, \sigma)$, $b_0 = 0$, $b_i = \frac{i}{I+1}$ $i = 1, \ldots, I+1$, $\sigma_i = \frac{f(b_i)-f(b_{i-1})}{b_i-b_{i-1}}$ $i = 1, \ldots, I+1$, and $f(b_i): b_i \to (b_i)^2$.

$$\sum_{g|g \neq g^*}(\gamma_g)^2 = \sum_{g \in G\{g^*\}} \left(\frac{n_g^1}{\overline{n}_g^1}\right)^2 - 2\left(\frac{\hat{n}_g^1}{\overline{n}_g^1}\right) n_g^1 + \left(\frac{\hat{n}_g^1}{\overline{n}_g^1}\right)^2 \quad (6)$$

In total there is $|G| - 1$ piecewise linear functions. The term $n_g^1/\overline{n}_g^1$ lies in the range [0,1] and as such, so does the term $\left(n_g^1/\overline{n}_g^1\right)^2$. The squared term is well approximated by a piecewise linear function over that interval and requires a relatively small number of intervals and breakpoints.

There are other alternatives to the sum of squares approach. Enforcing a more equitable change across all types is one such option. The following linear objective may instead be considered:

$$\text{Minimize } Z = \sum_{g \in G}(\gamma_g) \quad (7)$$

This variation is hereby denoted CMA-LIN. This approach still favours some types and not others and is not necessarily equitable, but it may be reasonable in some circumstances. Another alternative is to explicitly force an equitable increase or decrease (EQ-OBJ). The following objective and constraint can be imposed to achieve this:

$$\text{Minimize } Z = sgn(\delta) \times \lambda \text{ where } \gamma_g = \lambda \ \forall g \in G | g \neq g^* \quad (8)$$

The scaling previously discussed is vital for using an objective function like (8). The equality $\gamma_g = \lambda \ \forall g \in G | g \neq g^*$ is strict and that might be an issue sometimes. For instance, a particular patient type may not be able to be increased or decreased at all, as the required resources may already be saturated, or the lower or upper bound has already been reached. In response, we can instead enforce $\gamma_g \leq \lambda$, and add the following constraint:

$$n_g^1 \geq \hat{n}_g^1 - sgn(\delta) \times \lambda \overline{n}_g^1 \ \forall g \in G | g \neq g^* \quad (9)$$

This variation is hereby denoted CMA-EQ. It is worth noting that $\lambda.sgn(\delta) \leq \hat{n}_g^1/\overline{n}_g^1 \ \forall g \in G$ and a bound can be computed for $\lambda$.

**Final Remarks:** The model when solved may report no changes to the case mix. However, the output of the model is still useful. For instance, the model produces a plan of how to process the new case mix. A different allocation of resources may be necessary to achieve the alteration. The model may not be feasible in some circumstances, and that is informative of situations where an equitable alteration is not possible. An inequitable alteration however may be possible, but that requires a more sophisticated multicriteria approach and some additional direction and guidance about how to regulate competition between the different patient types.

**Demonstrative Example**. Let us consider a scenario of reduced size to demonstrate the case mix alteration models. In this scenario there are five intensive care beds, ten operating theatres, five recovery wards with (2, 5, 10, 14, 3) beds respectively, and five patient types. A feasible case mix of 113.53 patients (i.e., weekly) with $n^1$ =(5.68, 48.82, 20.43, 10.22, 28.38) determined from a prior analysis is to be altered. The case mix proportions are (0.05, 0.43, 0.18, 0.09, 0.25) and the patient types have the parameters shown in Table 1. Their names have been removed for confidentiality.

**Table 1.** Patient type parameters

| Type | $\overline{n}^1$ | # Sub Types | Sub Type | Sub Type Mix (%) | (icu, theatre, ward) time (# hrs) | Wards Used |
|---|---|---|---|---|---|---|
| T1 | 25.184 | 2 | T1-1 | 70 | (0, 1.2, 17.86) | Ward 1 |
|  |  |  | T1-2 | 30 | (6, 1.25, 8.35) | Ward 1, Ward 2 |
| T2 | 89.792 | 1 | T2-1 | 100 | (0, 2.4, 16.31) | Ward 1, Ward 2, Ward 5 |
| T3 | 65.477 | 3 | T3-1 | 25 | (0, 6.5, 12.94) | Ward 3 |
|  |  |  | T3-2 | 40 | (0, 4.56, 12.39) | Ward 3 |
|  |  |  | T3-3 | 35 | (0, 7.6, 5.54) | Ward 3 |
| T4 | 105.047 | 1 | T4-1 | 100 | (0, 3.4, 18.99) | Ward 4 |
| T5 | 70 | 1 | T5-1 | 100 | (12, 4.1, 22.81) | Ward 4, Ward 5 |

Table 2-5 describe the effect of various alterations and the solution of the variant CMA models. The results are rounded to 2 decimal places. The change to a particular patient type is shown in column two, and the change to all other patient types is shown in column four. In column six, the total corresponding change in the other patient types is shown. Depending on the method we also show $\lambda$ and $Z$. The selected changes provide a sufficient picture and for brevity concentrate upon describing what would happen if a patient type was eliminated completely or increased to the upper bound. Some other intermediate values are also considered.

From the CMA-EQ numerical testing in Table 2, it is worth noting a few things. It is possible, that some requested changes reduce all other patient types to zero. Also, if a patient type is increased too greatly, then some patient types would need to be reduced below zero to achieve a uniform equitable decrease. This makes no sense, however, and the model is not solvable in those circumstances. For example, when patient type T5 is increased to a level greater than 57.6, then it is impossible to achieve a uniform decrease. To obtain a solution, it is necessary to add in constraint (9). Hence, patient type T4 is decreased as much as possible, while the others are decreased equitably. The gamma value for patient type T4 is 0.097289 for all changes in patient type T5 above 57.6. As shown, when the increase is 57.9, then lambda is 0.0982 and 0.0982×105.047 = 10.3156 > 10.22.

In Table 4 and 5, the original CMA-SSQ approach is demonstrated. In Table 4 the results were obtained via quadratic programming, however, in Table 5 separable programming. The number of breakpoints used to approximate the squared term in the expansion of $(\gamma_g)^2$ was 500. There are subtle differences between the results of QP and SP, however, for the most part, only minor differences have been observed in decimal place accuracy.

The CMA-SSQ provides a more equitable case mix than CMA-LIN, but less equitable than CMA-EQ. Hence, it lies between the CMA-EQ and CMA-LIN approaches. This conclusion is justified by the values presented in Column 7 in Tables 2 - 4. The total impact shown in Column 7 is the sum of the alterations in Column 4, excluding those to the selected patient type.

**Table 2.** Demonstration of CMA-EQ approach for the selected case mix

| $g$ | Considered Revision | $\delta$ | Alterations | New Case Mix | $\mathbb{N}$ | Total Impact | $\lambda$ |
|---|---|---|---|---|---|---|---|
| 1 | 5.68 → 0 | -5.68 | (-5.68, 0.42, 0.48, 0.49, 0.33) | (0, 49.24, 20.91, 10.71, 28.71) | 109.57 | 1.71 | 0.0 |
| 1 | 5.68 → 2 | -3.68 | (-3.68, 0.27, 0.31, 0.32, 0.21) | (2, 49.09, 20.74, 10.54, 28.59) | 110.96 | 1.11 | 0.0 |
| 1 | 5.68 → 25 | 19.32 | (19.32, -1.41, -1.63, -1.65, -1.10) | (25, 47.41, 18.8, 8.57, 27.28) | 127.06 | -5.79 | 0.02 |
| 1 | 5.68 → 25.18 | 19.50 | (19.50, -1.43, -1.64, -1.67, -1.11) | (25.18, 47.39, 18.79, 8.55, 27.27) | 127.18 | -5.85 | 0.02 |
| 2 | 48.82 → 0 | -48.82 | (2.26, -48.82, 9.28, 9.42, 6.28) | (7.94, 0, 29.71, 19.64, 34.66) | 91.95 | 27.23 | 0.09 |
| 2 | 48.82 → 30 | -18.82 | (0.87, -18.82, 3.58, 3.63, 2.42) | (6.55, 30, 24.01, 13.85, 30.8) | 105.21 | 10.5 | 0.04 |
| 2 | 48.82 → 65 | 16.18 | (-0.75, 16.18, -3.07, -3.12, -2.08) | (4.93, 65, 17.36, 7.1, 26.3) | 120.69 | -9.02 | 0.03 |
| 2 | 48.82 → 89.79 | 40.97 | (-1.89, 40.97, -7.78, -7.9, -5.27) | (3.78, 89.79, 12.65, 2.32, 23.11) | 131.65 | -22.84 | 0.08 |
| 3 | 20.43 → 0 | -20.43 | (3.53, 12.59, -20.43, 14.73, 9.82) | (9.21, 61.41, 0, 24.95, 38.2) | 133.77 | 40.67 | 0.14 |
| 3 | 20.43 → 10 | -10.43 | (1.80, 6.43, -10.43, 7.52, 5.01) | (7.48, 55.25, 10, 17.74, 33.39) | 123.86 | 20.76 | 0.07 |
| 3 | 20.43 → 30 | 9.57 | (-1.65, -5.9, 9.57, -6.9, -4.6) | (4.03, 42.93, 30, 3.32, 23.79) | 104.07 | -19.04 | 0.07 |
| 3 | 20.43 → 65 | 44.57 | (-5.68, -47.61, 44.57, -10.22, -28.38) | (0, 1.215, 65, 0, 0) | 66.22 | -91.89 | 0.53 |
| 4 | 10.22 → 0 | -10.22 | (0.75, 2.68, 3.09, -10.22, 2.09) | (6.43, 51.5, 23.52, 0, 30.47) | 111.92 | 8.61 | 0.03 |
| 4 | 10.22 → 3 | -7.22 | (0.53, 1.89, 2.18, -7.22, 1.48) | (6.21, 50.71, 22.61, 3, 29.86) | 112.39 | 6.08 | 0.02 |
| 4 | 10.22 → 15 | 4.78 | (-0.35, -1.25, -1.44, 4.78, -0.98) | (5.33, 47.57, 18.99, 15, 27.41) | 114.3 | -4.02 | 0.01 |
| 4 | 10.22 → 105.05 | 94.83 | (-5.68, -34.07, -20.43, 94.83, -26.56) | (0, 14.75, 0, 105.05, 1.82) | 121.62 | -86.74 | 0.38 |

| g | Considered Revision | δ | Alterations | New Case Mix | ℕ | Total Impact | Z |
|---|---|---|---|---|---|---|---|
| 5 | 28.38 → 0 | -28.38 | (2.37, 8.46, 9.75, 9.9, -28.38) | (8.05, 57.28, 30.18, 20.12, 0) | 115.63 | 30.48 | 0.09 |
| 5 | 28.38 → 22 | -6.38 | (0.53, 1.90, 2.19, 2.26, -6.38) | (6.21, 50.72, 22.62, 12.48, 22) | 114.03 | 6.89 | 0.02 |
| 5 | 28.38 → 32 | 3.62 | (-0.30, -1.08, -1.24, -1.26, 3.62) | (5.38, 47.74, 19.19, 8.96, 32) | 113.27 | -3.88 | 0.01 |
| 5 | 28.38 → 57 | 28.62 | (-2.39, -8.53, -9.83, -9.98, 28.62) | (3.29, 40.29, 10.6, 0.24, 57) | 111.42 | -30.72 | 0.1 |
| 5 | 28.38 → 57.9 | 29.52 | (-2.47, -8.82, -10.16, -10.22, 29.52) | (3.21, 40, 10.27, 0,57.9) | 111.38 | -31.69 | 0.1 |
| 5 | 28.38 → 65 | 36.62 | (-3.31, -11.8, -13.59, -10.22, 36.62) | (2.37, 37.02, 6.84, 0, 65) | 111.23 | -75.54 | 0.13 |
| 5 | 28.38 → 70 | 41.62 | (-5.68, -1.74, -20.43, -10.22, 41.62) | (0, 47.08, 0, 0, 70) | 117.08 | -38.07 | 0.23 |

**Table 3.** Demonstration of CMA-LIN approach for the selected case mix

| g | Considered Revision | δ | Alterations | New Case Mix | ℕ | Total Impact | Z |
|---|---|---|---|---|---|---|---|
| 1 | 5.68 → 0 | -5.68 | (-5.68, 2.88, 0, 0, 0) | (0, 51.70, 20.43, 10.22, 28.38) | 110.73 | 2.88 | 0.03 |
| 1 | 5.68 → 2 | -3.68 | (-3.68, 1.88, 0, 0, 0) | (2, 50.69, 20.43, 10.22, 28.38) | 111.72 | 1.88 | 0.02 |
| 1 | 5.68 → 25 | 19.32 | (19.32, 0, -3.84, 0, 0) | (25, 48.82, 16.59, 10.22, 28.38) | 129.01 | -3.84 | 0.04 |
| 1 | 5.68 → 25.18 | 19.50 | (19.50, 0, -3.88, 0, 0) | (25.18, 48.82, 16.55, 10.22, 28.38) | 129.15 | -3.88 | 0.04 |
| 2 | 48.82 → 0 | -48.82 | (19.50, -48.82, 0, 0, 22.8) | (25.18, 0, 20.43, 10.22, 51.18) | 107.01 | 42.30 | 1.10 |
| 2 | 48.82 → 30 | -18.82 | (19.50, -18.82, 0, 0, 5.24) | (25.18, 30, 20.43, 10.22, 33.62) | 119.45 | 24.74 | 0.85 |
| 2 | 48.82 → 65 | 16.18 | (0, 16.18, -6.35, 0, 0) | (5.68, 65, 16.08, 10.22, 28.38) | 125.36 | -6.35 | 0.06 |
| 2 | 48.82 →89.79 | 40.97 | (0, 40.97, -16.09, 0, 0) | (5.68, 89.79, 4.34, 10.22, 28.28) | 138.31 | -16.09 | 0.16 |
| 3 | 20.43 → 0 | -20.43 | (10.5, 31.01, -20.43, 0, 6.51) | (25.18, 79.83, 0, 10.22, 34.89) | 150.12 | 48.02 | 1.21 |
| 3 | 20.43 → 10 | -10.43 | (19.50, 16.68, -10.43, 0, 0) | (25.18, 65.5, 10, 10.22, 28.38) | 139.28 | 36.19 | 0.96 |
| 3 | 20.43 → 30 | 9.57 | (0, 0, 9.57, -10.22, -5.78) | (5.68, 48.82, 30, 0, 22.6) | 107.1 | -16 | 0.18 |
| 3 | 20.43 → 65 | 44.57 | (-3.28, -48.82, 44.57, -10.22, -28.38) | (2.4, 0, 65, 0, 0) | 67.4 | -90.7 | 1.18 |
| 4 | 10.22 → 0 | -10.22 | (19.50, 4.61, 0, -10.22, 0) | (25.18, 53.43, 20.43, 0, 28.38) | 127.42 | 24.12 | 0.83 |
| 4 | 10.22 → 3 | -7.22 | (19.50, 0.36, 0, -7.22, 0) | (25.18, 49.18, 20.43, 3, 28.38) | 126.17 | 19.87 | 0.78 |
| 4 | 10.22 → 15 | 4.78 | (0, 0, -2.66, 4.78, 0) | (5.68, 48.82, 17.77, 15, 28.38) | 115.65 | -2.66 | 0.03 |
| 4 | 10.22 →105.05 | 94.83 | (0, -33.85, -20.43, 94.83, -28.38) | (5.68, 14.96, 0, 105.05, 0) | 125.69 | -82.66 | 0.98 |
| 5 | 28.38 → 0 | -28.38 | (19.50, 31.01, 0, 5.37, -28.38) | (25.18, 79.83, 20.43, 15.59, 0) | 141.03 | 55.88 | 1.17 |
| 5 | 28.38 → 22 | -6.38 | (19.50, 1.03, 0, 0, -6.38) | (25.18, 49.85, 20.43, 10.22, 22) | 127.68 | 20.54 | 0.79 |
| 5 | 28.38 → 32 | 3.62 | (0, 0, -2.43, 0, 3.62) | (5.68, 48.82, 18, 10.22, 32) | 114.72 | -2.43 | 0.02 |
| 5 | 28.38 → 57 | 28.62 | (0, 0, -19.21, 0, 28.62) | (5.68, 48.82, 1.23, 10.22, 57) | 122.95 | -19.21 | 0.19 |
| 5 | 28.38 → 57.9 | 29.52 | (0, 0, -19.81, 0, 29.52) | (5.68, 48.82, 0.62, 10.22, 57.9) | 123.24 | -19.81 | 0.19 |
| 5 | 28.38 → 65 | 36.62 | (0, 0, -20.43, -7.45, 36.62) | (5.48, 48.82, 0, 2.78, 65) | 122.08 | -27.88 | 0.27 |
| 5 | 28.38 → 70 | 41.62 | (-5.68, -1.74, -20.43, -10.22, 41.62) | (0, 47.08, 0, 0, 70) | 117.08 | -38.07 | 0.54 |

In Table 3, the alternative CMA-LIN approach is demonstrated. The resulting case mixes are quite different to those exhibited in Table 2. In general, the CMA-LIN tends to limit the number of patient types exhibiting a change. The CMA-EQ is more equitable but realises a bigger alteration than the CMA-LIN, resulting in less patients treated overall. Hence, the CMA-LIN produces case mixes with a greater number of patients.

**Table 4.** Demonstration of CMA-SSQ (QP) approach for the selected case mix

| g | Considered Revision | δ | Alterations | New Case Mix | ℕ | Total Impact | Z |
|---|---|---|---|---|---|---|---|
| 1 | 5.68 → 25 | 19.32 | (19.32, -0.69, -2.34, -1.34, -0.72) | (25, 48.13, 18.09, 8.89, 27. 66) | 100.11 | -5.09 | 0.0 |
| 1 | 5.68 → 25.18 | 19.50 | (19.50, -0.7, -2.36, -1.35, -0.73) | (25.18, 48.12, 18.07, 8.86, 27.66) | 127.89 | -5.14 | 0.0 |
| 2 | 48.82 → 65 | 16.18 | (-0.05, 16.18, -4.16, -2.39, -1.28) | (5.63, 65, 16.27, 7.83, 27.10) | 121.83 | -7.88 | 0.0 |
| 2 | 48.82 → 89.79 | 40.97 | (-0.12, 40.97, -10.53, -6.04, -3.24) | (5.56, 89.79, 9.9, 4.18, 25.14) | 134.57 | -19.93 | 0.02 |
| 3 | 20.43 → 30 | 9.57 | (-0.18, -4.4, 9.57, -8.52, -4.56) | (5.51, 44.43, 30, 1.7, 23.82) | 105.46 | -17.65 | 0.01 |
| 3 | 20.43 → 65 | 44.57 | (-3.28, -48.82, 44.57, -10.22, -28.38) | (2.4, 0, 65, 0, 0) | 67.4 | -90.7 | 0.49 |
| 4 | 10.22 → 15 | 4.78 | (-0.02, -0.59, -2, 4.78, -0.62) | (5.66, 48.23, 18.42, 15, 27.76) | 115.07 | -3.234 | 0 |
| 4 | 10.22 → 105.05 | 94.83 | (-1.32, -33.18, -20.43, 94.83, -28.38) | (4.36, 15.64, 0, 105.05, 0) | 125.05 | -83.31 | 0.34 |
| 5 | 28.38 → 32 | 3.62 | (-0.02, -0.5. -1.69, -0.97, 3.62) | (5.66, 48.32, 18.74, 9.25, 32) | 113.97 | -3.18 | 0 |
| 5 | 28.38 → 57 | 28.62 | (-0.16, -3.95, -13.36, -7.66, 28.62) | (5.23, 44.87, 7.07, 2.56, 57) | 116.73 | -25.13 | 0.02 |
| 5 | 28.38 → 57.9 | 29.52 | (-0.16, -4.08, -13.78, -7.9, 29.52) | (5.52, 44.74, 6.65, 2.32, 57.9) | 117.13 | -25.92 | 0.03 |
| 5 | 28.38 → 65 | 36.62 | (-0.2, -5.06, -17.09, -9.81, 36.62) | (5.48, 43.76, 3.34, 0.41, 65) | 117.99 | -32.16 | 0.04 |
| 5 | 28.38 → 70 | 41.62 | (-5.68, -5.6, -18.91, -10.22, 41.62) | (0, 43.22, 1.52, 0, 70) | 114.74 | -40.41 | 0.1 |

**Table 5.** Demonstration of CMA-SSQ (SP) approach for the selected case mix

| g | Considered Revision | δ | Alterations | New Case Mix | ℕ | Total Impact | Z |
|---|---|---|---|---|---|---|---|
| 1 | 5.68 → 0 | -5.68 | (-5.68, 2.88, 0, 0, 0) | (0, 51.70, 20.43, 10.22, 28.38) | 110.73 | 2.88 | 0.0 |
| 1 | 5.68 → 2 | -3.68 | (-3.68, 1.87, 0, 0, 0) | (2, 50.69, 20.43, 10.22, 28.38) | 111.72 | 1.87 | 0 |
| 1 | 5.68 → 25 | 19.324 | (19.32, -0.72, -2.39, -1.26, -0.7) | (25, 48.1, 18.04, 8.96, 27.68) | 127.78 | -5.07 | 0.0 |
| 1 | 5.68 → 25.18 | 19.504 | (19.50, -0.72, -2.43, -1.26, -0.7) | (25.18, 48.1, 18, 8.96, 27.68) | 127.92 | -5.11 | 0.0 |
| 2 | 48.82 → 0 | -48.82 | (19.50, -48.82, 0, 0, 22.80) | (25.18, 0, 20.43, 10.22, 51.18) | 107.01 | 42.3 | 0.71 |
| 2 | 48.82 → 30 | -18.82 | (19.50, -18.82, 0, 0, 5.24) | (25.18, 30, 20.43, 10.22, 33.62) | 119.45 | 24.74 | 0.61 |
| 2 | 48.82 → 65 | 16.18 | (-0.05, 16.18, -4.13, -2.46, -1.26) | (5.63, 65, 16.3, 7.76, 27.12) | 121.81 | -7.9 | 0.0 |
| 2 | 48.82 → 89.79 | 40.972 | (-0.151, 40.97, -10.52, -6.08, -3.21) | (5.53, 89.79, 9.91, 4.14, 25.17) | 134.54 | -19.96 | 0.02 |
| 3 | 20.43 → 0 | -20.43 | (19.50, 31.01, -20.43, 0, 6.51) | (25.18, 79.83, 0, 10.22, 34.89) | 150.12 | 57.02 | 0.73 |
| 3 | 20.43 → 10 | -10.43 | (19.50, 16.68, -10.43, 0, 0) | (25.18, 65.5, 10, 10.22, 28.38) | 139.28 | 36.18 | 0.63 |
| 3 | 20.43 → 30 | 9.57 | (-0.2, -4.48, 9.57, -8.4, -4.61) | (5.48, 44.34, 30, 1.83, 23.77) | 105.42 | -17.69 | 0.01 |
| 3 | 20.43 → 65 | 44.57 | (-3.28, -48.82, 44.57, -10.22, -28.38) | (2.4, 0, 65, 0, 0) | 67.4 | -90.7 | 0.49 |
| 4 | 10.22 → 0 | -10.22 | (19.50, 4.61, 0, -10.22, 0) | (25.184, 53.43, 20.43, 0, 28.38) | 127.424 | 24.11 | 0.6 |
| 4 | 10.22 → 3 | -7.22 | (19.50, 0, 0, -7.22, 0.21) | (25.184, 48.82, 20.43, 3, 28.59) | 126.024 | 19.71 | 0.6 |
| 4 | 10.22 → 15 | 4.78 | (-0.05, -0.54, -2.06, 4.78, -0.56) | (5.63, 48.28, 18.37, 15, 27.82) | 115.1 | -3.21 | 0.0 |
| 4 | 10.22 → 105.05 | 94.827 | (-1.31, -33.18, -20.43, 94.83, -28.38) | (4.37, 15.64, 0, 105, 0) | 125.01 | -83.3 | 0.33 |
| 5 | 28.38 → 0 | -28.38 | (19.50, 31.01, 0, 5.37, -28.38) | (25.18, 79.83, 20.43, 15.59, 0) | 141.03 | 55.88 | 0.72 |
| 5 | 28.38 → 22 | -6.38 | (19.50, 1.03, 0, 0, -6.38) | (25.18, 49.85, 20.43, 10.22, 22) | 127.68 | 20.53 | 0.6 |
| 5 | 28.38 → 32 | 3.62 | (-0.05, -0.54, -1.65, -0.99, 3.62) | (5.63, 48.28, 18.78, 9.23, 32) | 113.92 | -3.23 | 0 |
| 5 | 28.38 → 57 | 28.62 | (-0.15, -3.94, -13.42, -7.55, 28.62) | (5.53, 44.88, 7.01, 2.67, 57) | 117.09 | -25.06 | 0.02 |
| 5 | 28.38 → 57.9 | 29.52 | (-0.15, -4.12, -13.73, -7.97, 29.52) | (5.53, 44.7, 6.71, 2.25, 57.9) | 117.09 | -25.97 | 0.03 |
| 5 | 28.38 → 65 | 36.62 | (-0.2, -5.02, -17.08, -9.86, 36.62) | (5.48, 43.8, 3.35, 0.37, 65) | 118 | -32.16 | 0.04 |
| 5 | 28.38 → 70 | 41.62 | (-5.68, -5.56, -18.93, -10.22, 41.62) | (0, 43.26, 1.5, 0, 70) | 114.76 | -40.39 | 0.1 |

**Further Discussion**: A change to the main case mix was discussed above. When applying the proposed CMA models, the patient sub mixes implied by the original starting case mix are maintained. Whatever proportions are inherent therein, are not altered. In many situations we would expect the impact to other patient types to be of the same order of magnitude of the original alteration, but inverse to it. However, some alterations of the case mix permit higher or lower increases (decreases) to be realised. These impacts are not directly proportional to the scale of the original alteration. This means that there may be some latent unused capacity, relative to the original case mix.

Specific changes to a particular patient sub type are also worth considering. A similar process and model can be posed for this situation. The exact details are shown below:

Step 1. Select a current case mix $n^2$, and designate targets $\hat{n}^2_{g,p} = n^2_{g,p}$.

Step 2. Select a change of $\delta$ units in one patient sub type $(g^*, p^*)$, i.e., $(\delta, g^*, p^*) | -\hat{n}^2_{g^*,p^*} \leq \delta \leq \overline{n}^2_{g^*,p^*} - \hat{n}^2_{g^*,p^*}$ where $\delta \neq 0$ and $\overline{n}^2_{g,p}$ is the upper bound on the number of patients of sub type $(g, p)$ that are treatable.

Step 3. Solve the CMA variant model below to obtain the case mix $n^2$. Report the differences $\gamma_{g,p}$.

$$\text{Minimize } sgn(\delta) \times Z$$
$$\text{where } Z = \sum_{g \in G} \sum_{p \in P_g} (\gamma_{g,p})^2 \text{ or } Z = \sum_{g \in G} \sum_{p \in P_g} \gamma_{g,p} \text{ or } Z = \lambda \tag{10}$$

Subject to:
$$n^2_{g^*,p^*} = \hat{n}^2_{g^*,p^*} + \delta \tag{11}$$
$$\gamma_{g^*,p^*} = 0 \text{ and } \gamma_{g,p} = sgn(\delta) \times \left(\frac{\hat{n}^2_{g,p} - n^2_{g,p}}{\overline{n}^2_{g,p}}\right) \forall g \in G, \forall p \in P_g | g \neq g^*, p \neq p^* \tag{12}$$
$$sgn(\delta) \times n^2_{g,p} \leq sgn(\delta) \times \hat{n}^2_{g,p} \forall g \in G, \forall p \in P_g | g \neq g^*, p \neq p^* \tag{13}$$
$$n^2_{g,p} \geq \hat{n}^2_{g,p} - sgn(\delta) \times \lambda \overline{n}^2_{g,p} \forall g \in G, \forall p \in P_g | g \neq g^*, p \neq p^* \text{ (CMA-EQ only)} \tag{14}$$
$$\gamma_{g,p} \geq 0 \ \forall g \in G, \forall p \in P_g \tag{15}$$
$$n^2_{g,p} \geq \mu^2_{g,p} n^1_g \ \forall g \in G, \forall p \in P_g | g \neq g^* \text{ (Sub mix adherence)} \tag{16}$$
+ From Appendix A, the constraints (A2) - (A8)

In constraint (16) it is necessary to point out that sub mix proportions are only enforced for unaffected patient types, i.e., $g \in G | g \neq g^*$.

**Demonstrative Example.** Consider the patient sub type cohort (a.k.a., mix) ([3.97, 1.7], [48.82], [5.11, 8.17, 7.15], [10.22], [28.38]) associated with the previously considered case mix. If sub type T1-1 was chosen and an alteration of 5 was selected, then $\hat{n}^2_{1,1} \rightarrow 8.97$. For the CMA-EQ option the following sub mix is obtained, $n^2 =$((8.97, 1.32), (48.54), (4.99, 7.99, 6.99), (9.9), (28.16)). Associated with that sub mix are the following alterations, ((5,-0.38),(-0.277),(-0.12.-0.18,-0.16),(-0.32),(-0.22)). Hence, the new case mix is $n^1 =$(10.29, 48.54, 19.96, 9.9, 28.16) and $\mathbb{N} = 116.85$.

## 2.3. Case Mix Comparisons

In this section, multi-criteria decision-making theory is adapted to help support hospital CMP. A multicriteria decision support system (MCDSS) is also proposed.

In hospital CMP, the identification of a single case mix solution that meets certain conditions is important and has been predominantly sought in the methods proposed in previous articles. The identification of a few alternative case mix solutions is valuable and constructive. As shown in Section 2.2, further analysis of a single case mix solution is worthy. It also makes sense to start with a single case mix solution and to suggest alterations. In the previous section, we considered how to assess the impact of any change. A mathematical model was proposed for that purpose. It produces a second case mix and provides the decision maker with two case mixes to compare. However, this raises the question, how does a decision maker determine which is preferable?

In multi-criteria CMP, it is essential to identify an assortment of potential solutions, namely a Pareto frontier of alternatively optimal case mix solutions. As there are multiple conflicting objectives, no single solution exists that simultaneously optimizes each objective. Pareto optimality is a useful concept. It is worth noting that all Pareto optimal solutions are considered equally good without further preference information. A solution is called Non-Dominated, Pareto Optimal, or Pareto efficient, if none of the objective functions can be improved without degrading some other objective values. To obtain a Pareto frontier, there are various techniques, the basis of which is the solution of an underlying mathematical optimization model with multi-objectives.

In Burdett and Kozan (2016), a 21 objective CMP problem was solved, and tens of thousands of Pareto optimal case mix solutions were found. From a practical perspective, it is unclear how those solutions would be accessed and compared, and how a single solution could be chosen by hospital planner, managers, and executives. The first step to handling this dilemma seems to be to provide a means of scoring an individual case mix solution, and second to provide a means of critiquing two case mix, providing insight into which is better or worse. This aligns well with the task discussed in Section 2.2. To the best of our knowledge previous articles have not formally scored patient case mix solutions, nor provided techniques to compare competing case mix solutions purely on the basis of number of patients treated. Other considerations are abundant but are outside the scope of this article.

### 2.3.1. Scoring Case Mix

Every case mix $n^1$ can be compared to the ideal and anti-ideal case mix. The ideal case mix denoted $\overline{n}^1$ occurs when each patient type achieves its' upper bound. The anti-ideal is the case mix with zero patients of each type, which is hereby denoted $\underline{n}^1$. The proximity of $n^1$ to $\overline{n}^1$ is an important indicator of progress for obvious reasons. The proximity can be computed according to various metrics, and the following is most sensible.

$$proximity = 100 \times \frac{\|n^1 - \overline{n}^1\|_2}{\|\overline{n}^1 - \underline{n}^1\|_2} \text{ where } \|v^1 - v^2\|_2 = \left(\sum_g \frac{(v_g^1 - v_g^2)^2}{(\epsilon_g)^2}\right)^{1/2} \quad (17)$$

The proximity value lies in the range [0,100], and takes the value zero at the utopia point (i.e., when $n^1 = \overline{n}^1$). The denominator is the distance between the ideal and anti-ideal, and the numerator is the distance between the ideal and the current case mix. As some patient types may have vastly different orders of magnitude, the proximities should be scaled, to maintain objectivity. The value $\epsilon_g$ is a user defined measure of significance for patients of type $g$, and the default value is $\epsilon_g = \overline{n}_g^1$. Finally, we can compute progress as $100 - proximity$. The proximity of two case mix can also be calculated using equation (17).

**Demonstrative Example:** If we compare case mix (5.68, 48.82, 20.43, 10.22, 28.38) and (16.46, 71.67, 11.79, 10.59, 24.39) where the ideal solution is (25.18, 89.79, 65.48, 105.05, 70), the diagram in Figure 1 is useful and may be produced. It shows scaled proximity information. Evidently, the second case mix is further from the anti-ideal and closest to the ideal. The second case mix has higher numbers of patient type T1 and T2, and similar numbers of T4 and T5. Only in patient type T3 is it more noticeably deficient. The scaled distance of 0.518 indicates some difference, but the extent of the dissimilarity is not sufficiently clear. All we can say is that there is an aggregated squared difference of $0.518^2=0.269$.

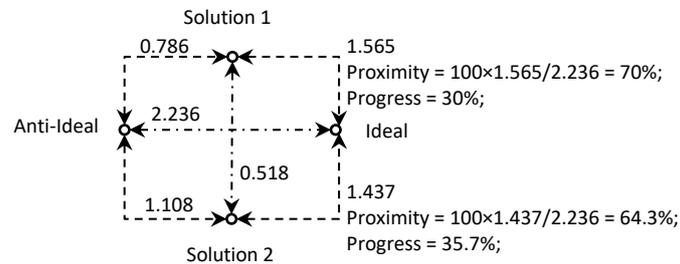

**Figure 1.** Comparing case mix solutions and their proximity

**Final Remarks**. Scoring case mix by proximity to the ideal is a useful metric. In equation (17) other norms may also be used, like the 1-norm. The significance level is inherently subjective, and highly influential in the assessment. Two case mix may have the same proximity yet be placed in a totally different parts of the objective space. In those circumstances, the approach is not useful for making any type of judgement. It seems that an iterative approach, involving an alteration of the level of significance is necessary, to build up sufficient evidence, of the merit of one case mix over another. It is necessary to identify if small changes to the level of significance change the resulting comparison, and how those changes skew the proximity.

### 2.3.2. Quantifying Similarity and Dissimilarity

As discussed, proximity is not a full proof concept to judge achievement or case mix similarity and dissimilarity. Definition 1 and its' corollary are self-evident and may be used instead, as a more rigorous way to perform a comparison. Definition 1 is based upon the e-dominance principles of Laumanns et. al. (2002), later adapted for instance by Hancock et. al. (2015). When using the concept of e-dominance, the user is expected to provide a value that represents the minimum amount of change in an objective that is considered significant. In other words, the user defines what a significant difference is. In some scenarios, a difference of one unit implies significant difference, but in others, it would not. The application of the e-dominance principle is akin to partitioning the objective space into regions. The approach maintains diversity equally in all regions of the Pareto frontier, regardless of where significant trade-off is or is not occurring.

**Definition 1:** Given two case mix solutions $n^{1(A)}$ and $n^{1(B)}$, where $n^{1(A)}, n^{1(B)} \subset \mathbb{R}^{|G|}$, we can say that the solutions are similar (i.e., they are not significantly different) if the distance $\left| n_g^{1(A)} - n_g^{1(B)} \right| \leq$ $\epsilon_g \ \forall g \in G$. Otherwise, they are not similar, and significant trade-offs must be realised if one solution is selected over another.

**Corollary**: If $\left|n_g^{1(A)} - n_g^{1(B)}\right| \leq \epsilon_g$ then the change in patient type $g$ is not significant. If $\left|n_g^{1(A)} - n_g^{1(B)}\right| > \epsilon_g$ then we can conclude significant difference. The level of similarity (LOS) is $100 \times \left(\frac{1}{|G|}\right) \sum_{g \in G \mid \left|n_g^{1(A)} - n_g^{1(B)}\right| \leq \epsilon_g}$ (1) and the level of dissimilarity (LOD) is one minus the LOS.

**Corollary**: Around any solution $n^1$, the boundary of the region of similarity-dissimilarity can be identified, for instance by identifying all solutions $n^{1(*)}$ such that $\epsilon_g < \left|n_g^1 - n_g^{1(*)}\right| \leq \lambda \epsilon_g \ \forall g \in G$.

**Demonstrative Example:** If we compare case mix (5.68, 48.82, 20.43, 10.22, 28.38) and (16.46, 71.67, 11.79, 10.59, 24.39) and define $\epsilon = (2.5, 9, 6.5, 10.5, 7)$ then, only in patient type T4 and T5 is there a lack of significant difference. Hence, we could say that the level of similarity 40% (i.e., 2/5) and the level of dissimilarity is 60%. These numbers can be added to Figure 1 on the arc connecting the two case mixes.

It is worth noting that around case mix one, the boundary of similarity is as follows, ([3.18, 8.18], [39.82, 57.82], [13.93, 26.93], [0, 20.7], [21.38, 35.38]), when $\lambda = 1$.

**2.3.3. Comparing Case Mix**

It is desirable to make a judgement upon which case mix, say $n^{1(A)}$ or $n^{1(B)}$, is most preferable, if not better or worse. To the best of our knowledge, no such approach has yet been provided in the literature. Determining the exact nature of the trade-offs that will be incurred is also needed. To do that, we propose Definition 2. The concept behind Definition 2 is to find the net effect as shown in Figure 2. In essence, we propose that differences between the case mixes are partitioned into gains and losses. The gains and losses can then be aggregated using the concept of a resultant vector. This produces two values which can be easily compared, to make a judgement. Compared to an approach involving $|G|$ comparisons the new approach is more transparent and user friendly. Our approach is primarily defined to critique differences in the number of patients of each type, and to make judgements relative to that metric. Other metrics perhaps qualitative, however, can also be compared. Financial metrics, however, can be aggregated in to a single "dollar" value and may be less well suited to this type of analysis.

**Definition 2**: Given two case mix solutions $n^{1(A)}$ and $n^{1(B)}$, where $n^{1(A)}, n^{1(B)} \subset \mathbb{R}^{|G|}$, we can say that one solution is preferable if the ratio of the net gain to the net loss, namely $R = \mathcal{G}^+/\mathcal{G}^-$, where $\mathcal{G}^+ = \|V^+\|_2$ and $\mathcal{G}^- = \|V^-\|_2$, is not roughly one. In other words, the next gain is not roughly the same as the net loss. Here, $V^+ = \sum_{g \in G \mid \delta_g > 0} \delta_g \tilde{u}_g$ is the resultant "gain" vector and $V^- = \sum_{g \in G \mid \delta_g < 0} \delta_g \tilde{u}_g$ is the resultant "loss" vector, where $\tilde{u}_g$ are "unit" basis vectors and $\delta_g = \hat{n}_g^{1(B)} - \hat{n}_g^{1(A)}$ or $\delta_g = \left(n_g^{1(B)} - n_g^{1(A)}\right)/\epsilon_g$ and $\hat{n}_g^1 = (n_g^1 - \underline{n}_g^1)/(\overline{n}_g^1 - \underline{n}_g^1)$. When using $\delta_g = (n_g^{1(B)} - n_g^{1(A)})/\epsilon_g$ we can say that one solution is "significantly" better.

**Corollary**: If $\mathcal{G}^+ < \mathcal{G}^-$ then $n_g^{1(A)} \succcurlyeq n_g^{1(B)}$ where "$\succcurlyeq$" is used to signify "better than". Similarly, if $\mathcal{G}^+ > \mathcal{G}^-$ then $n_g^{1(B)} \succcurlyeq n_g^{1(A)}$.

**3D Example**: Consider case mix (1,20,16) and (10,5,35) where $x \in [0,15]$, $y \in [0,30]$, $z \in [3,50]$. The normalized points are (0.066, 0.666, 0.32) and (0.666, 0.166, 0.7). The net differences are (+9,-15,+19) and (+0.6,-0.5,+0.38) after normalization. The net gain is 0.71 and the net loss is 0.5. Hence, case mix 2 is preferable. The net effect is not directly towards the ideal, but to the right of it.

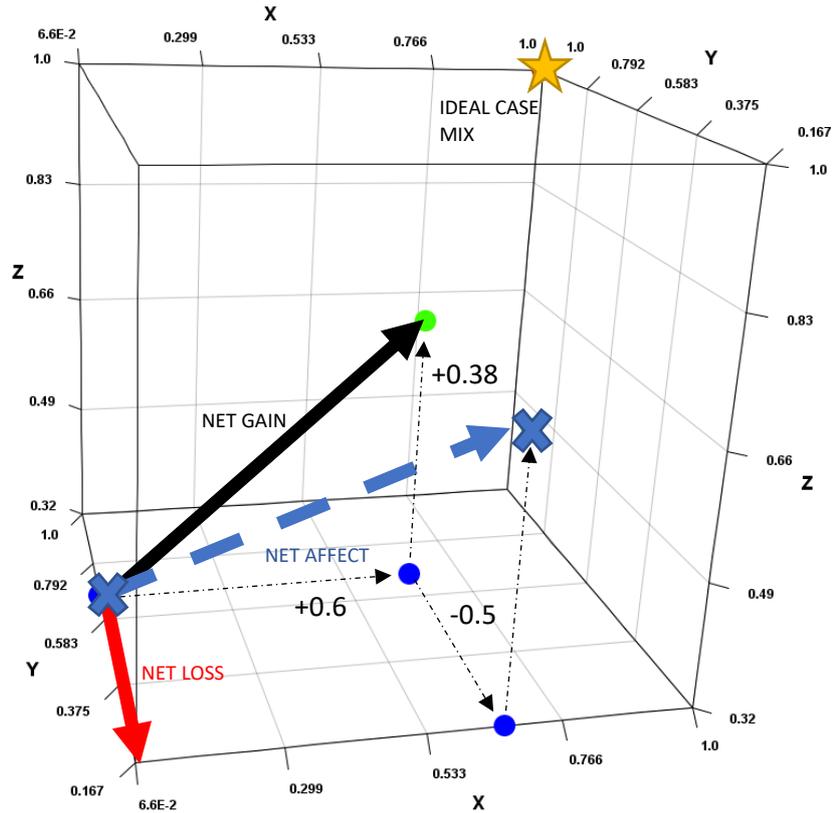

**Figure 2**. Comparing case mix (1,20,16) and (10,5,35)

**Higher Dimensional Example**: When comparing case mix one (5.68, 48.82, 20.43, 10.22, 28.38) to case mix two (16.46, 71.67, 11.79, 10.59, 24.39), we note that case mix 2 has 21.37 more patients. There is a direct gain of 34 patients (i.e., spread across $g = 1, 2$ and 4), and a loss of 12.63 (i.e., spread across $g = 3$ and 5). The differences are as follows, (10.78, 22.85, -8.64, 0.37, -3.99). As such, $V^+ = (0.43, 0.25, 0, 0.0032, 0)$ and $V^- = (0, 0, 0.132, 0, 0.057)$. Hence, the net gain is $\mathcal{G}^+ = 0.498$ and the net loss is $\mathcal{G}^- = 0.144$ with a ratio of $R = 3.465$. The second case mix is preferable as the net gain exceeds the net loss.

It is foreseeable that a decision maker may also be interested in trade-offs occurring between some patient types and not others, which are deemed unimportant for one reason or another. The proposed approach can easily be modified to assess net gains and losses only in those types, as opposed to the automatic inclusion of all patient types. An assessment where multiple comparisons are performed separately to identify if comparatively higher amounts of deterioration in one or more patient types occurs is also possible.

**Demonstrative Example**: Reconsidering the previous case mixes let us assume that trade-offs between patient type T3, T4 and T5 are important. The considered differences are (-8.64, 0.37, -3.99). As such, $V^+ = (0, 0.00352, 0)$ and $V^- = (0.132, 0, 0.057)$. Hence, the net gain is $\mathcal{G}^+ = 0.00352$ and the net loss is $\mathcal{G}^- = 0.144$ with a ratio of $R = 0.245$. Therefore, the first case mix is preferable as the net loss exceeds the net gain.

**Final Remarks**. The case mix comparisons described relate only to $n^1$. For a particular $g \in G$, different sub mixes can also be compared. Specifically, Definition 2 can be applied to compare sub mix $n_g^{2(A)}$ and $n_g^{2(B)}$ where $n_g^{2(.)} = \left(n_{g,1}^2, n_{g,2}^{2(.)}, \ldots, n_{g,|P_g|}^{2(.)}\right)$. Two different sub mixes can be compared in entirety as well. For instance, it is possible to compare $\text{vec}(n^{2(A)})$ and $\text{vec}(n^{2(B)})$. There are however many more criterions to compare (i.e., $\sum_{g \in G} P_g$).

## 3. Putting Theory into Practice

To facilitate hospital case mix planning activities, a prototype PDST was created in Burdett et. al. (2022b). The tool permits users to determine the maximum number of patients that may be treated over time, subject to case mix and time availability constraints. This task is performed by solving the CMA model shown in Appendix A. Given user defined targets, a best fit case mix is also obtainable, by solving a non-linear variant of the CMA model. The feasibility of a specified case mix can also be checked within the PDST.

The two tasks discussed in previous sections, namely case mix alteration and case mix comparison have been added to the PDST and suitable graphical user interfaces (GUI) have been constructed. The details of the new extensions are discussed in due course. Both tasks however rely upon the upper bound (a.k.a., limit) $\hat{n}_g^1$ for each patient type. This bound describes the maximum number of patients of type $g$ that can be realised if the resources of the entire hospital were used to process only that patient type.

The window shown in Figure 3 is provided for users to activate this assessment. The "Bound Analysis" button activates the solution process, involving the solution of the CMA model $|G|$ times. During iteration $g \in G$, the case mix is set as $\mu_g^1 = 1$ and $\mu_{g'}^1 = 0 \ \forall g' \in G | g' \neq g$. The results shown in the right pane are generated and displayed progressively, instead of all at once.

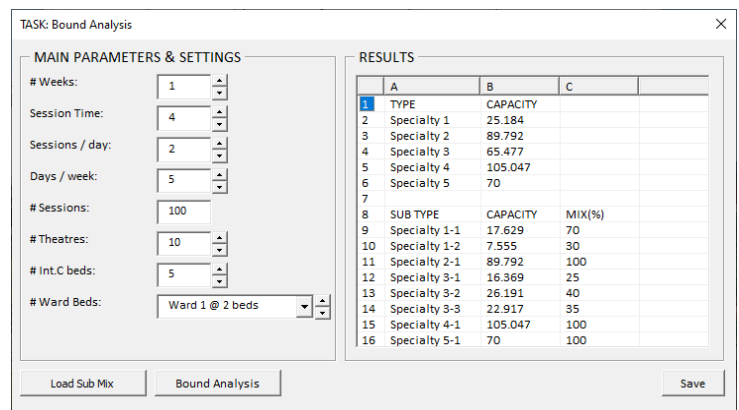

**Figure 3**. GUI to perform a bound analysis

### 3.1. Case Mix Alteration and Impact Assessments

To alter a case mix and to analyse the effect, the GUI shown in Figure 4 has been added. The first step is to specify the current case mix using the Load Cohort" button. The combo box summarises the different patient types, their current $n_g^1$ value, plus for reference purposes, the upper bound $\overline{n}_g^1$. Any patient type can be selected and altered, using the drop-down selection mechanism. The "Analyse Change" button informs the user to enter a $-n_g^1 \leq \delta \leq \overline{n}_g^1$ value, and if satisfactory, the CMA model is then solved. In the bottom ListView the results are then displayed. The before and after values are shown plus the required change. The user is asked whether the results are accepted or rejected. Further alterations to other patient types are then permitted to be analysed. In other words, a hierarchical assessment is facilitated.

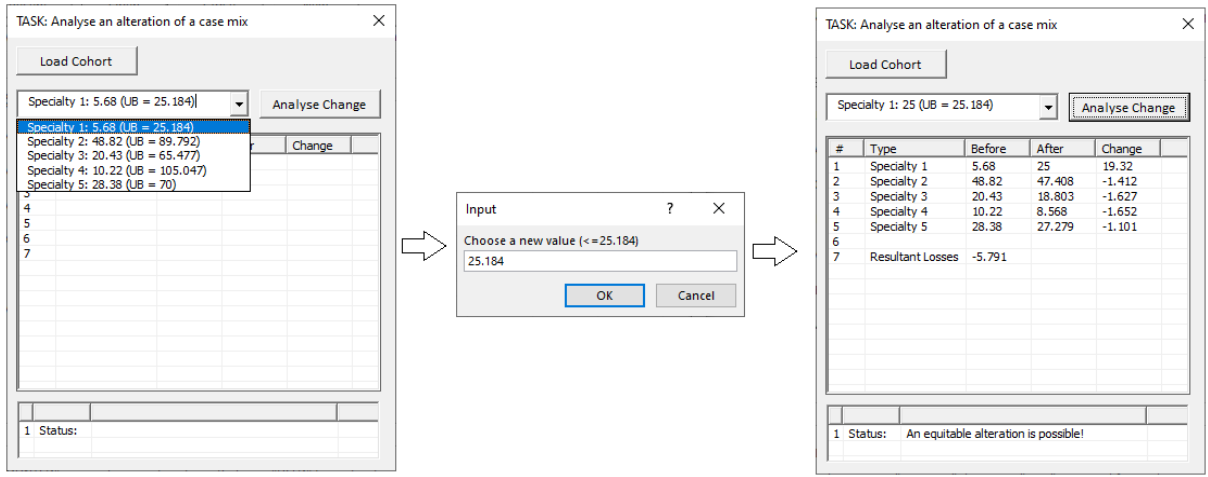

**Figure 4.** GUI to facilitate case mix alteration

Figure 5 shows an alternative approach. The bar chart in Figure 5 is essentially a set of sliders. Each slider is for a specific patient type. The current value for each patient type is shown as marker on a bar, whose height is relative to the upper bound. A single patient type can then be selected easily and incremented and decremented to whatever value the user likes. The change to other patients can then be shown immediately, and the overall effect can be witnessed. For instance, changes in the height of each marker can be visualised. The man downside of this idea is that the number of bars may be difficult to see and manipulate for scenarios with hundreds of patient types.

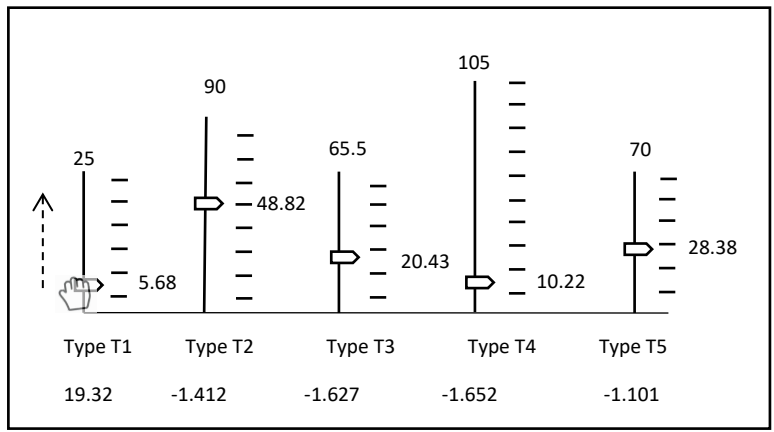

**Figure 5.** An alternative interface (of sliders) for users to manipulate

### 3.2. Case Mix Comparisons

To critique two case mix, the GUI displayed in Figure 6 has been developed. It is worth noting that SQ is an abbreviation for a squared value and SC for a scaled value. Also, Min refers to the nadir solution and Max the ideal. In that GUI the first step is to load two candidate case-mix. The next step is to choose a level of significance for each patient type. If none is selected, then the present upper bounds are used. The "Compare" button activates the assessment, and applies the calculations from Section 2.3, and specifically those associated with Definition 2. The GUI currently shows all calculations involved. This highly detailed output is for experts and is extraneous to the average end-user. In future versions, this information could possibly be hidden, accessible only in reports or by direct query.

    Figure 6 shows a situation where a level of significance has been defined. In column H, differences exceeding $\epsilon_g$ are highlighted using (*). The level of significance that has been used, has amplified the gains and losses, and placed both case mixes closer to the ideal.

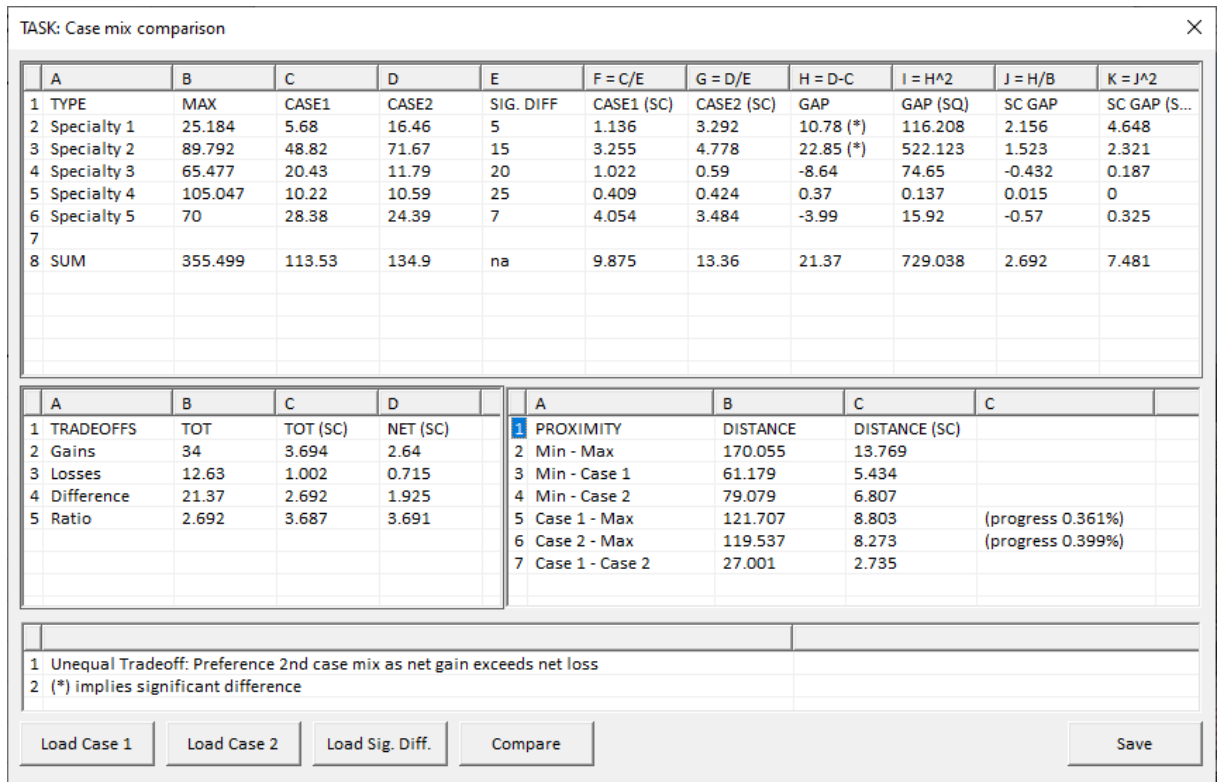

**Figure 6.** GUI to facilitate case mix comparisons

## 4. COVID Inspired Case Study

Covid-19 is an infectious disease (a.k.a., coronavirus) that causes mild to moderate respiratory illness in most people. Originating at the end of 2019 it has been "in play" for about two years. In the early stages many countries around the world were unable or unwilling to quarantine this virus and consequently its' spread throughout the world has been assured. To the best of our knowledge further impacts and misery are expected for many more years, for instance as mutations occur and produce new variants of the virus, and as the virus progresses through the population.

It is safe to say that there would be very few hospitals in the world that have not been seriously impacted by covid-19. Around the world the virus has interfered with the day-to-day operations of most. For instance, this virus has caused many additional representations, and many of those patients have been seriously ill. To handle that demand, many elective surgeries been postponed, and many wards and beds have been repurposed. Intensive care facilities, used for ventilating patients mechanically, have been particularly stretched.

This articles' methods are well suited to analysing the impact of covid-19 or any other virus or medical disaster on the capacity of a hospital. To demonstrate that assertion let us reconsider the earlier scenario with five intensive care beds, ten operating theatres, five recovery wards and five patient types. Let us now consider a longer time frame of two months and the arrival of additional COVID patients who are admitted to the hospital for treatment. We will now identity the effect on the current case mix of 113.53 patients per week where $n^1$ =(5.68, 48.82, 20.43, 10.22, 28.38). The covid-19 patients are defined as a sixth patient type, namely T6. Amongst that cohort, various sub types can be defined. We define four sub types based upon Gulsen (2020), according to severity of illness at presentation. The clinical pathway and average length of stay for each sub type in show in Table 6. These values were adapted from the data summarised in Vekaria et. al. (2021) and Whitfield et. al. (2020). To cope with anticipated demand during the pandemic, a new ward has been set up (i.e., Ward 6) and several current wards (i.e., Ward 1 and 5) have been repurposed as shown in Table 7. All covid patients will be kept in COVID specific wards to restrict transmission.

**Table 6.** New patient type and sub types

| Sub Type | Mix (%) | Summary | Pathway | Length of stay |
|---|---|---|---|---|
| T6-1 (mild) | 45 | Patient with mild upper respiratory tract infection | Ward | 0.25 days in a ward |
| T6-2 (moderate) | 35 | Patients requiring hospitalization, with pneumonia and with/without the need for oxygen | Ward | 5 days in a ward |
| T6-3 (severe) | 15 | Patients who need ICU treatment and require non-invasive or invasive mechanical ventilation, or with acute respiratory distress and/or non-pulmonary involvement | (Ward, ICU, Ward) | 2 days in ward prior to ICU, 5 days in ICU + 7 days in ward |
| T6-4 (critical) | 5 | Patients who need immunomodulatory therapy or with multi-organ failure and/or cytokine storm | (ICU, Ward) | 14 days in ICU + 7 days in ward |

**Table 7.** Revised hospital layout

| Ward | Repurposed for covid | #Beds | Used For |
|---|---|---|---|
| Ward 1 | ✓ | 2 | T6 |
| Ward 2 | ✗ | 5 | T1-1, T1-2, T2-1 |
| Ward 3 | ✗ | 10 | T3-1, T3-2, T3-3 |
| Ward 4 | ✗ | 14 | T4-1, T5-1, T1-1, T1-2, T2-1 |
| Ward 5 | ✓ | 3 | T6 |
| Ward 6 | ✓ | 6 | T6 |

**Table 8.** New patient type information and bounds

| Type | $\overline{n}^1$ (orig) | $\overline{n}^1$ (new) | # Sub Types | Sub Type | Sub Type Mix (%) | (icu, theatre, ward) time (# hrs) | Wards Used |
|---|---|---|---|---|---|---|---|
| 1 | 201.47 | 1000 | 2 | 1-1 | 70 | (0, 1.2, 17.86) | Ward 2, Ward 4 |
|   |        |      |   | 1-2 | 30 | (6, 1.25, 8.35) | Ward 2, Ward 4 |
| 2 | 718.33 | 1000 | 1 | 2-1 | 100 | (0, 2.4, 16.31) | Ward 2, Ward 4 |
| 3 | 523.82 | 523.82 | 3 | 3-1 | 25 | (0, 6.5, 12.94) | Ward 3 |
|   |        |        |   | 3-2 | 40 | (0, 4.56, 12.39) | Ward 3 |
|   |        |        |   | 3-3 | 35 | (0, 7.6, 5.54) | Ward 3 |
| 4 | 840.38 | 840.38 | 1 | 4-1 | 100 | (0, 3.4, 18.99) | Ward 4 |
| 5 | 560 | 560 | 1 | 5-1 | 100 | (12, 4.1, 22.81) | Ward 4 |
| 6 | na | 172.91 | 4 | 6-1 | 45 | (0,0,6) | Ward 1, Ward 5, Ward 6 |
|   |    |        |   | 6-2 | 35 | (0,0,120) | Ward 1, Ward 5, Ward 6 |
|   |    |        |   | 6-3 | 15 | (120,0,216) | Ward 1, Ward 5, Ward 6 |
|   |    |        |   | 6-4 | 5 | (336,0,168) | Ward 1, Ward 5, Ward 6 |

It is worth noting that we have added surgery times to ward length of stay for this analysis, as beds are acquired before surgery begins. Subject to the proportional case mix (0.05, 0.43, 0.18, 0.09, 0.25, 0) the treatable cohort (i.e., hospital capacity) is 908 patients of the following types $n^{1(\text{orig})}$ =(45.41, 390.54, 163.48, 81.74, 227.06, 0). After the hospital's layout is changed, the capacity to treat non covid patients is not altered, even though fewer wards and beds are available. This occurs because some restrictions have been relaxed. For instance, Ward 4 has been permitted to treat more patient types.

Let us now consider case mix alterations required as a result of treating covid patients of type T6. Let us also consider whether the hospital reconfiguration is sufficient to meet the demand for covid patients, and whether further wards should be repurposed. Pre-analysis shows that no more than 21.5 covid-19 patients can be treated per week, given the sub mix (0.45, 0.35, 0.15, 0.5). Hence, we analyse the effect of T6 patients in the range (0, 172].

**Table 9.** Application of CMA-EQ approach

| $\delta$ | Alterations | New Case Mix ($n^1$) $\mathbb{N}$ ($\mathbb{N} - \delta$) | Ward Util (%) (ICU, OT, W1, W2, W3, W4, W5, W6) |
|---|---|---|---|
| 50 | (-0.0025, -0.0026, -0.001, -0.002, -0.001, 100) | (45.41, 390.54, 163.48, 81.74, 227.06, 50) 958.22 (908.22) | (67.66, 100, 0, 9.02, 19.75, 81.73, 0, 53.01) |
| 100 | (-0.0025, -0.0026, -0.001, -0.002, -0.001, 100) | (45.41, 390.53, 163.48, 81.74, 227.06, 100) 1008.2 (908.22) | (93.55, 100, 100, 0, 19.75, 84.95, 100, 22.69) |
| 125 | (-45.41, -52.79, -27.65, -44.36, -29.56, 125) | (0, 337.75, 135.83, 37.38, 197.5, 125) 833.46 (708.46) | (100, 80.54, 100, 0, 16.41, 66.28, 100, 49.2) |
| 150 | (-45.41, -182.25, -95.47, -81.74, -102.06, 150) | (0, 208.29, 68.01, 0, 125, 150) 551.3 (401.3) | (100, 44.62, 100, 0, 8.22, 35.59, 100, 75.71) |
| 172 | (-45.41, -296.18, -155.14, -81.74, -165.86, 172) | (0, 94.36, 8.34, 0, 61.2, 172) 335.9 (163.9) | (100, 16.51, 100, 0, 1, 18.14, 100, 99.03) |

Table 9 shows that, the hospital can meet covid-19 patient demands quite easily, up to a point, roughly 12-13 patients/week, and has some capacity to spare. As the number increases further, the original cohort of different patient types is greatly affected and maintaining an equitable decrease reduces overall outputs considerably. However, some of the original cohort, can still be treated. The exact number is shown in brackets in column 3.

**Table 10.** Application of CMA-LIN approach

| $\delta$ | Alterations | New Case Mix ($n^1$) $\mathbb{N}$ ($\mathbb{N} - \delta$) | Ward Util (%) (ICU, OT, W1, W2, W3, W4, W5, W6) |
|---|---|---|---|
| 50 | (0, 0, -0.005, 0, 0, 50) | (45.41, 390.54, 163.48, 81.74, 227.06, 50) 958.22 (908.22) | (67.66, 100, 0, 9.02, 19.75, 81.73, 0, 53.01) |
| 100 | (0, 0, -0.005, 0, 0, 100) | (45.41, 390.53, 163.48, 81.74, 227.06, 100) 1008.2 (908.22) | (93.55, 100, 100, 0, 19.75, 84.95, 44.2, 50.6) |
| 125 | (0, 0, 0, 0, -36.37, 125) | (45.41, 390.54, 163.48, 81.74, 190.69, 125) 996.86 (871.86) | (100, 95.34, 100, 0, 19.75, 79.75, 89.58, 54.41) |
| 150 | (0, 0, 0, 0, -108.87, 150) | (45.41, 390.54, 163.48, 81.74, 118.19, 150) 949.36 (799.36) | (100, 86.05, 100, 0, 19.75, 69.38, 100, 75.71) |
| 172 | (0, 0, 0, 0, -172.67, 172) | (45.41, 390.54, 163.48, 81.74, 54.39, 172) 907.56 (735.56) | (100, 77.88, 100, 0, 19.75, 60.25, 100, 99.03) |

Table 10, however shows that if equity does not matter, then the hospital can still treat a high number of patients, relative to "pre-covid" times. However, some patient types can be exploited, like T5. The model found that reducing T5 admissions, provides the capacity to treat most of the original cohort and the new covid patients.

**Table 11.** Application of CMA-SSQ(QP) approach

| $\delta$ | Alterations | **New Case Mix ($n^1$)** $\mathbb{N}$ ($\mathbb{N} - \delta$) | Ward Util (%) (ICU, OT, W1, W2, W3, W4, W5, W6) |
|---|---|---|---|
| 50 | (-0.01, -0.01, -0.01, -0.01, -0.01, 50) | (45.41, 390.54, 163.48, 81.74, 227.06, 50) 958.17 (908.17) | (67.66, 100, 0, 0, 19.75, 84.94, 0, 53.01) |
| 100 | (-0.01, -0.01, -0.01, -0.01, -0.01, 100) | (45.41, 390.53, 163.48, 81.74, 227.06, 100) 1008.2 (908.22) | (93.55, 100, 66.77, 0, 19.75, 84.95, 62.67, 52.43) |
| 125 | (-16.23, -0.012, -0.016, -0.014, -33.94, 125) | (29.18, 390.52, 163.46, 81.73, 193.12, 125) 983.01 (858.01) | (100, 95.03, 72.98, 0, 19.75, 78.69, 71.91, 72.25) |
| 150 | (-45.41, -0.012, -0.011, -0.011, -102.06, 150) | (0, 390.53, 163.47, 81.73, 125, 150) 910.73 (760.7) | (100, 85.2, 82.15, 0, 19.75, 66.44, 84.5, 89.41) |
| 172 | (-41.41, -0.014, -0.012, -0.013, -165.86, 172) | (0, 390.53, 163.47, 81.73, 61.12, 172) 868.92 (696.92) | (100, 77.02, 98.81, 0, 19.75, 57.31, 57.31, 99.29, 9.79) |

Table 11 shows the SSQ approach, and demonstrates the "in-between" approach, that is a little bit more equitable than the CMA-LIN variant. Types T1 and T5 are reduced the most.

**Final Remarks:** The above analysis could be repeated for any layout alteration that is being considered and for any sized hospital. An iterative approach considering a sequence of changes is also appropriate. In the above situation, if we later decided that more than 21.5 covid-19 patients were to be treated per week, another reconfiguration would have to be envisaged and analysed.

## 5. Conclusions

This article proposes analytical techniques to support hospital case mix planning. This emerging topic considers the ramifications of treating different patient cohorts and ultimately seeks to identify the type of patient cohort that should be treated, above all others. Ultimately, the choice of case mix is synonymous to choosing a single Pareto optimal solution in a multi-criteria objective space.

Though the idea of CMP is appealing, the reality is that most hospitals do not apply any formal CMP techniques. They operate dynamically and treat patients as they emerge, considering severity of illness to prioritise one patient type over another. The master surgical schedule is also influential, chosen to satisfy surgeons and their availability, at the expense of all other considerations.

To improve the state of the art, we consider how an alteration to an existing patient case mix either frees up capacity for other patient types or eliminates it. Our mathematical optimization model identifies the impact to other patient types and identifies how many extra patients of each type can be treated, or which patient types need to be reduced, and by how much. Our approach provides hospital planners a sensitivity analysis, that may be used to inform and guide an iterative approach, to obtain the most appealing case mix.

We also developed some techniques to assess, compare and critique competing case mix. A score based upon proximity to the ideal was first proposed. Measures of similarity and dissimilarity were then proposed. These are based upon the concept of e-dominance. Last, an approach to measure trade-offs and to aggregate those into a net gain and net loss were developed. These statistics permit an end-user a means of judging overall merit, and a way to judge which case mix is superior.

In summary, the proposed approach forces an end user to clarify and disclose their belief around the value of treating patient types in different numbers. And a means to adapt their beliefs or requirements. In this article, the number of patients of each type was treated as the main objective. However, other objectives may also be considered, like reimbursement or revenue. Regarding the uptake of these methods, graphical user interfaces were proposed, implemented, and tested. The resulting decision support tool looks viable, and further developments are being considered.

**Acknowledgements:** This research was funded by the Australian Research Council (ARC) Linkage Grant LP 180100542 and supported by the Princess Alexandra Hospital and the Queensland Children's Hospital in Brisbane, Australia.

**Appendix A. CMP Model**

$n_g^1$: Number of patients of type $g$
$n_{g,p}^2$: Number of patients of sub type $(g,p)$
$\alpha_{(g,p,k),w}$: Number of activities of type $(g,p,k)$ allocated to zone $w$

Maximize $\mathbb{N} = \sum_g n_g^1$  (Aggregated outputs) (A1)
Subject To:
$n_g^1 = \sum_{p \in P_g}(n_{g,p}^2) \quad \forall (g,p) \in GP$  (Aggregated sub types) (A2)
$n_{g,p}^2 = \sum_{w \in l_{(g,p,k)}} \alpha_{(g,p,k),w} \quad \forall (g,p) \in GP, \forall k \in \{1..|A_{g,p}|\}$ (Allocation of resource to activities) (A3)
$0 \leq n_g^1 \leq \overline{n}_g^1 \quad \forall g \in G$  (Permitted Range) (A4)
$0 \leq n_{g,p}^2 \leq \overline{n}_{g,p}^2 \quad \forall g \in G, \forall p \in P_g$ (Permitted Range) (A5)
$0 \leq \alpha_{(g,p,k),w} \leq \overline{n}_{g,p}^2 \quad \forall g \in G, \forall p \in P_g, \forall a = (g,p,k) \in A_{g,p}, \forall w \in l_{(g,p,k)}$  (Permitted Range) (A6)
$U_w = \sum_{a \in A_w} \alpha_{a,w} t_a \quad \forall w \in W$  (Utilization) (A7)
$U_w \leq b_w T_w \quad \forall w \in W$  (Restricted time availability) (A8)

$n_g^1 \geq \mu_g^1 \mathbb{N} \quad \forall g \in G$      (Case mix adherence)      (A9)

$n_{g,p}^2 \geq \mu_{g,p}^2 n_g^1 \quad \forall g \in G, \forall p \in P_g$    (Sub mix adherence)      (A10)

Constraints (A1) – (A3) are basic book-keeping equations. These equations model the inherent relationship between the number of patients of each type and sub type. Also, there is an inherent relationship between the sub types and the allocation of resources. The main decision variables have restricted ranges as shown in (A4) – (A6). Resource usage and saturation is restricted by constraint (A7) and (A8). Case mix and sub mix are enforced by (A9) and (A10).